# *Explaining the Logical Nature of Electrical Solitons in Neural Circuits*

by J. R. Burger

**Introduction**

Memory-based activities within a brain depend on neurons being able to generate arbitrary Boolean logic. A single neuron may be modeled as follows: Inputs may be assumed to be excitatory and inhibitory positive ions, commonly termed neurotransmitters, usually coupled to receivers located along dendrites; outputs are excitatory or inhibitory neurotransmitter ions from the tips of an axon. Dendritic receivers are negative inside relative to outside because of small charge transfers through the membrane. Thus positive neurotransmitter ions are attracted to negative receivers and can trigger an electrical pulse.

Solitons are, by definition, stable moving pulses that do not disperse as they travel along a conductor (dispersion is a widening of the width of a pulse). Solitons are noteworthy in many engineering disciplines. The soliton phenomenon was first described by (1808–1882) who observed a solitary wave in the Union Canal (a canal in Scotland). Optical solitons in optical fibers routinely transfer large amounts of information over thousands of kilometers. Electromagnetic solitons can be created with the help of PN junctions or ferromagnetic beads to send picosecond-width pulses through a transmission line without dispersion [1].

Electrical solitons in dendrites and axons rely not on nonlinear inductance or capacitance, but on an active membrane. Without reviewing charge transfer theories of the past, the thin neural membrane, with dimensions in the nanometer range, clearly enables charge transfer to and from a cell. Basic physics still applies. Neural membranes obviously are held tight by a strong electric field, since they support a rest voltage across an extremely short distance.

Membranes can be modeled in different ways, all giving a similar neural pulse. The author's favorite approach is to model the membrane as ferroelectric, a typical property of dielectrics such as manufactured purple membrane films. Neural membranes are thin enough to permit charge transfer, conjectured to be the tunneling result of thermally energetic ions and electrons. Triggering is instigated either by positive (excitatory neurotransmitter) ions or by positive charge within the region being triggered. Either way, triggering reduces the electric field locally and unlocks membrane particles.

The membrane, once unlocked, permits a transfer of negative charge from inside the neuron (alternately, positive charge appears inside). It may be assumed that all ions, including sodium and potassium, are involved in charge transfer once the membrane is triggered (there is no need for separate types of ion channels in this model). But since potassium is heavier and less effective for charge transfer, sodium is more important initially. As a result, current sources in the model are enabled at t = 0; a sodium related current source pumps current into the neuron; a potassium related current source pumps current out of the neuron, but at a lower rate. Within a millisecond or two, a positive voltage of a few tens of millivolts builds up in membrane capacitance, thus strongly reversing the electric field.

The reversed field pulls the membrane's ferroelectric particles into a new orientation. The local field of the rotated particles hinders the actions of external sodium ions, but not the actions of internal ions such as potassium. In the model below, the <u>sodium</u> controlled current source is switched off when voltage reaches a given level; internal voltage begins to drop.

Unless the membrane is re-triggered, internal voltage will undershoot the rest value by a few millivolts. Undershoot is terminated when the restored electric field forces the rotated membrane particles to return to normal, thus reducing charge transfers to near zero. In the model below, the <u>potassium</u> controlled current source is switched off at a specified undershoot value.

After undershoot, the following model allows a simple decay to negative rest voltage of -70 mV by way of membrane conductance.

Obviously there has to be a way to stop a neural action, because a neuron must quickly recover to be used again for many additional operations. It may be concluded that electrical solitons both instigate and terminate a neural action.

In the following ferroelectric model of a membrane, the membrane may be triggered by positive charge within the conductor. Also, as long as the internal voltage is below a certain threshold, the membrane may be retriggered. Retriggering is critical to the generation of reflected solitons. Reflected solitons are quite common in the model below and serve to encourage stray neurotransmitter ions to stay away from the neuron. The bulk of the electrostatic repelling, however, is accomplished by the first generated soliton of positive charge. Physics dictates that this positive charge push away positive ions clinging to the membrane.

To begin, there is only one soliton in the dendrites; however, once a soliton leaves the vicinity, assuming neurotransmitters have not yet experienced re-uptake, positive neurotransmitter ions may be re-attracted back to the receivers to trigger another soliton. A train of dendritic solitons, relatively wide-spaced in time, say 100 Hz, is theoretically possible. A variation of this may be occurring in the pyramidal neuron in which synapses occur on relatively distant receivers.

Pulse waveforms in dendrites may differ from waveforms in soma and axon, owing to differences in ionic compositions and membrane properties. A given soliton may have sufficient duration to trigger a short train of soma pulses in a burst, close-spaced in time, say 300 Hz as is commonly observed. Note, propagating solitons are quite common, but they are not the only way to trigger the soma. A short train of soma pulses, close-spaced in time, may occur in the Purkinje neuron because of synapses directly to the soma.

Inhibitory neurotransmitter ions may be assumed to surround a local region of a neural pathway; in fact, their synapses are observed on dendrites near the soma. They apparently serve to prevent triggering in this local region, turning the path into a passive conductor. Solitons do not repel inhibitory ions; in fact, inhibitory ions repel solitons. Note that inhibitory ions do not have to recycle back to their sources immediately, but rather may depend on thermal agitation for their eventual removal. In this case, the logic they generate will stay in place a little longer, usually not a problem in logical circuits.

The above overview of neural actions provides a framework for the following simulations.

**Simulations**

Simulations provide useful insight into electrical activity in neurons that otherwise is unavailable. Here we apply a basic WinSpice simulator. The simulations below assume that sodium and potassium currents switch on and off as a function of the voltage across a membrane.

The main parameters may be chosen based on published data and basic physical calculations. Rest potential is assumed to be -70 mV. Both sodium (Na) and potassium (K) currents are assumed triggered at -55 mV for simulation purposes. Table 1 lists major parameters.

Table 1
**Simulation Parameters**

| Rest Potential | -70 mV |
|---|---|
| Trigger Voltage (both Na and K currents) | -55 mV |
| Na Current Cutoff Voltage | +50 mV |
| K Current Cutoff Voltage | -95 mV |
| Sodium Current | 0.1345 mA/cm$^2$ |
| Potassium Current | 0.0608 mA/cm$^2$ |
| Membrane Capacitance | 1 uF/cm$^2$ |
| Membrane Conductance | 0.3 mA/cm$^2$ |
| Internal Resistivity | 15.7 Ω-cm |

Hand analysis of a single neural pulse results in Figure 1. The figure shows Na and K currents being triggered simultaneously. First Na cuts off at roughly 1.5 ms; then K cuts off at roughly 3.5 ms, allowing the voltage to relax to its equilibrium value of about -70 mV.

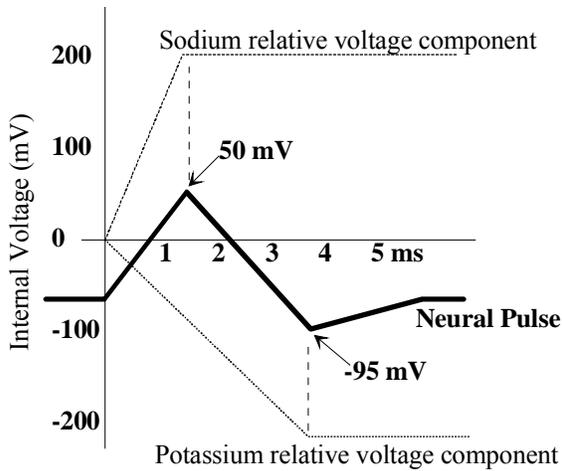

**Figure 1  Simplified analysis of a neural pulse in terms of Na and K currents.**

The switching properties of a neural membrane may be simulated using elementary nonlinear amplifiers to generate hysteresis loops.  Figure 2 shows the necessary control signals.

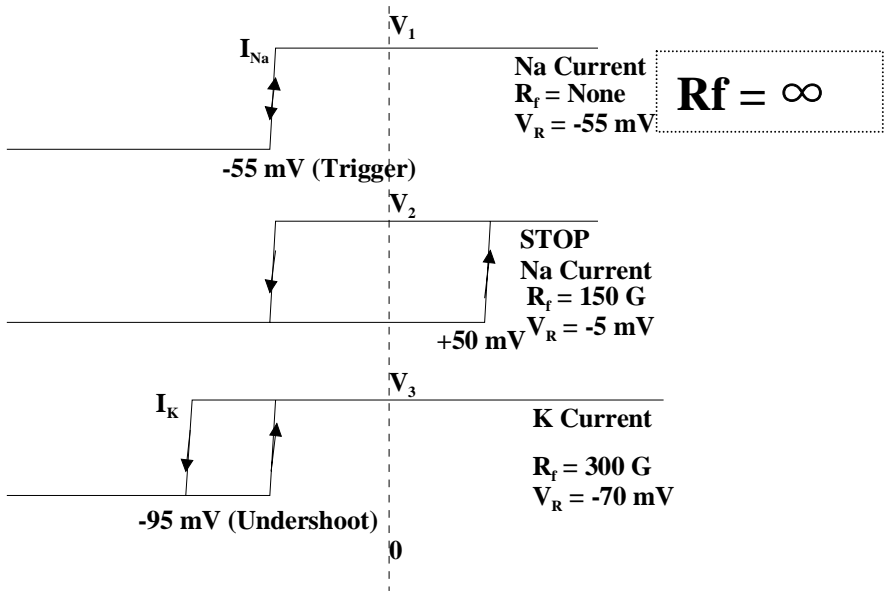

**Figure 2  Control signals to simulate membrane current switching.**

Triggering (of both Na and K currents) results when internal voltage exceeds -55 mV, sending $V_1$ high.  The simulation model cuts off the Na current at about +50 mV, as regulated by $V_2$, the STOP signal; re-triggering is impossible until the internal voltage drops below about -55 mV because of $V_2$.  Below a voltage of about -55 mV the pulse can always be re-triggered.  K current is stopped between at about -95mV as regulated by $V_3$.

Each control signal is generated by a subcircuit as in Figure 3.  The resistances involved in the voltage-controlled voltage sources are chosen in the giga ohm range to avoid loading the high resistances of the neural path. The circuit is initialized via 100 pF to force $V_i = 0$, for i = 1, 2, 3 at t = 0.

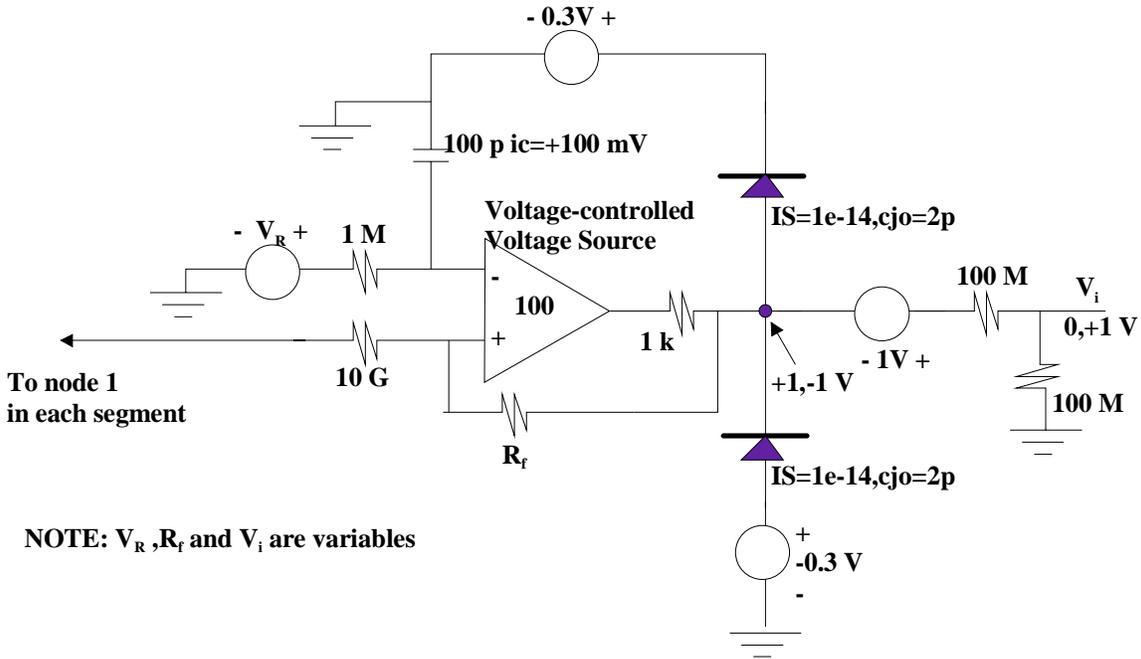

**Figure A3   Control subcircuit.**

The simulation approach is to model a neural path, axon or dendrite, as short segments that are connected in series. Series segments are illustrated in Figure 4.

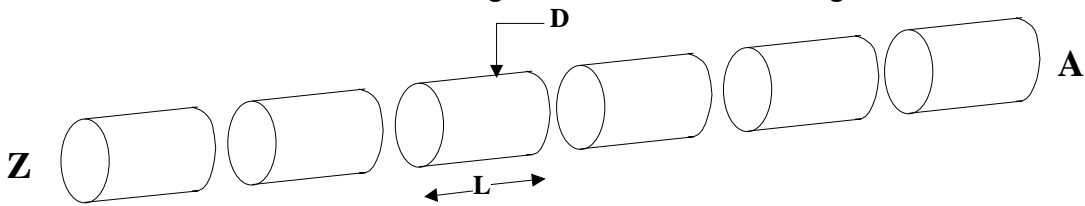

**Figure 4   Segments model of a neural path.**

This simulation used ten segments to study propagation from segment A to segment Z. Each segment has circuit elements as in Figure 5.

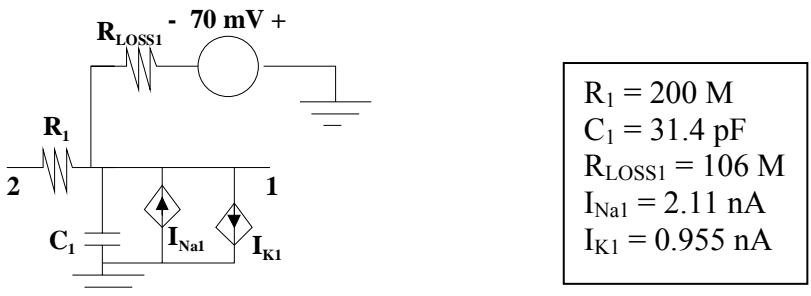

$R_1 = 200$ M
$C_1 = 31.4$ pF
$R_{LOSS1} = 106$ M
$I_{Na1} = 2.11$ nA
$I_{K1} = 0.955$ nA

**Figure 5   Circuit model of a segment.**

Note that the leakage conductance of a segment is used to model the charging attributed to the Nernst potential across the membrane. A segment can be scaled to any convenient size. In units of centimeters, microfarads and milliamperes, the parameters were calculated as follows:

$$L = 0.1 \text{ cm};\ D = 1.0\text{ e-4 cm};\ A_{CS} = \pi (D/2)^2;\ R_1 = 15.7\ L/A_{CS};\ A_{SIDE} = \pi DL;$$
$$C_1 = 1\ A_{SIDE};\ R_{LOSS1} = 1/(0.3\text{ e-3 }A_{SIDE});\ I_{Na1} = 0.1345\ A_{SIDE};\ I_K = 0.0608\ A_{SIDE}$$

The current sources are modeled with voltage-controlled current sources; $I_K$ is under the control of $V_3$. $I_{Na}$ is activated by $V_1$; but $V_1$ is soon zeroed by $V_2$ using a voltage-controlled switch. The circuit for this is given in Figure 6.

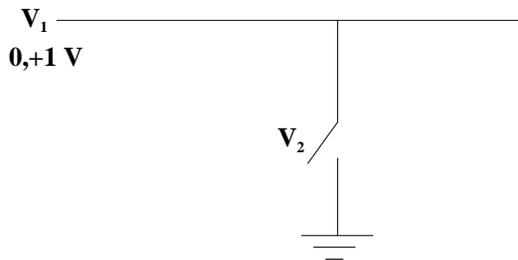

**Figure 6  Sodium current stop switch.**

Segment A was triggered by assuming a 10 nA current pulse for about 0.2 ms. The result was a propagating pulse as in Figure 7. The propagation of a pulse is quite interesting, as those who perform this simulation will discover; the result represents a component of an action potential. This is by definition a soliton, or a solitary pulse whose general waveform resists dissipation as the pulse propagates. There are no reflections in Figure 4 because the terminating segment does not have a capacitive load.

Each segment has a capacitance of about 31 pF and a resistance of about 200 M. Loss resistance is about 106 M. If an additional 60 pF is presented to the last segment only, with no change in resistance, there is a reflection as in Figure 8. This suggests that pulses arriving at the larger capacitance of a soma are reflected (back propagated) back down dendrites. This occurs even if the soma does not activate. Simulations with the above model indicate that reflections are quite common for a variety of discontinuities in the series resistance and shunt capacitance.

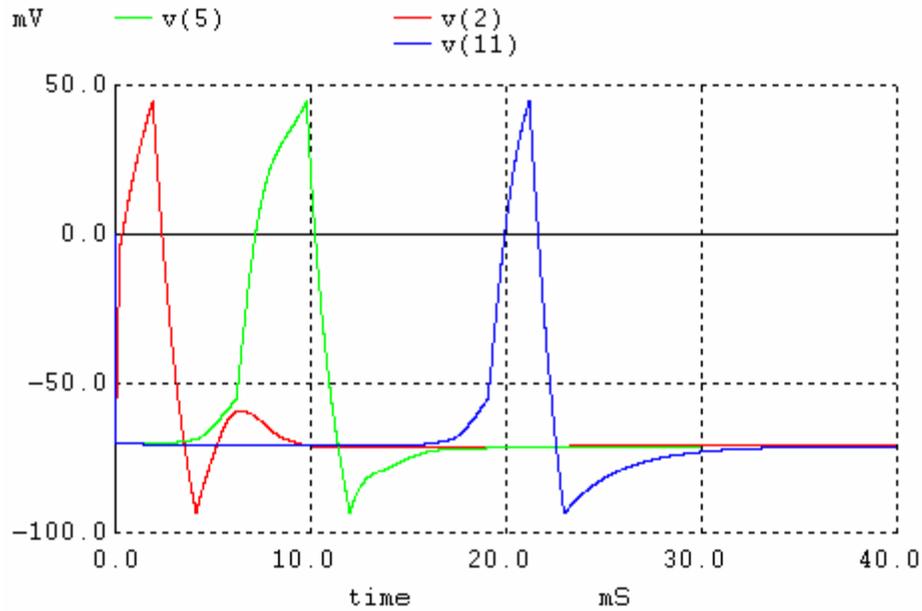

**Figure 7  Propagating Pulse Moving From Segment 2 to 5 to 11.**

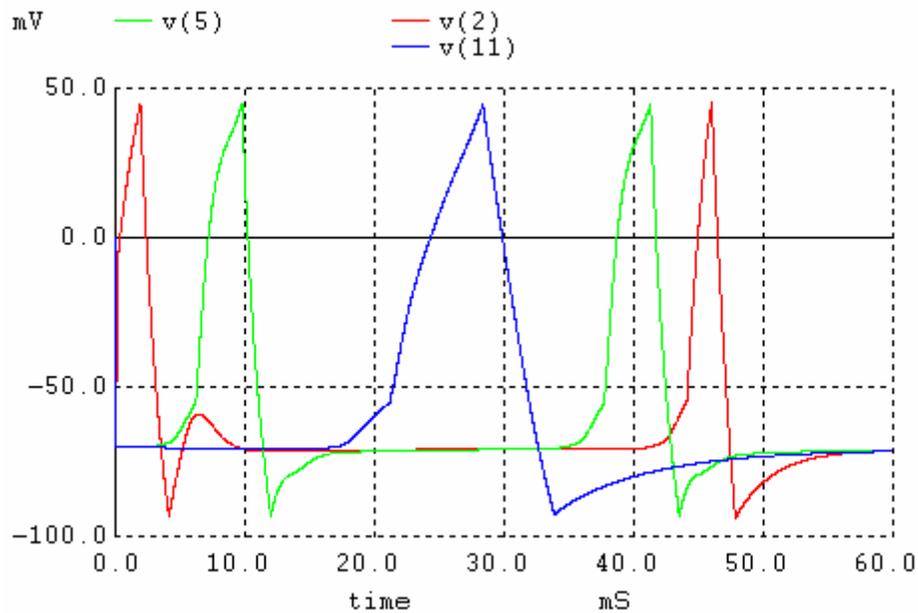

**Figure 8  Soliton that is reflected by a load capacitance of 60 pF.**

A neural system does not have to involve colliding solitons.  This is fortunate, since colliding solitons cannot pass through each other like electromagnetic solitons do. Typically, membrane supported solitons will annihilate as observed in Figure 9.  Note that just before the annihilation, the pulses become steeper.  Compare v(5) to v(6).  This is because the charges in the pulses are attracted to each other through the series resistance, and pull each other down, preventing a trigger to continue the propagation.

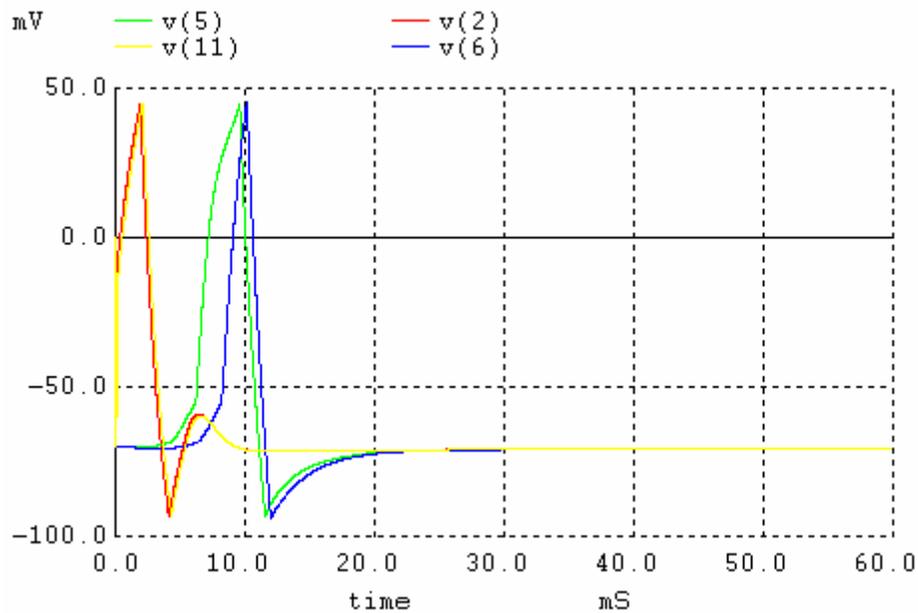

**Figure 9  Solitons injected at each end will collide and annihilate; no output V(11).**

## Logic generation in dendrites

Neural paths in dendrites are natural logic gates without having to rely on soma activation.  Two input paths beginning with segments A and B can be constructed as in Figure 10.  In the first example, all segments are assumed identical.  Segments 5 and 25 are joined to the input of segment 6.  First only segment A is triggered.  The result is a propagation that travels to the junction, splits, and travels to both B and Z as in Figure 11.  Z receives a pulse if both segments A and B are triggered (Figure 12), giving a form of the OR gate.

It may be noted in this model that if the capacitance at the input to segment 6 is reduced by about 33 %, the exclusive OR gate results.  A single pulse will go through, but two pulses will 'collide' and annihilate (Figure 13).  This sort of reduction in local capacitance could involve local myelination, although it could also results by a small increase in the geometry of the junction.  Given an XOR gate, the NOT function could be constructed without difficulty.

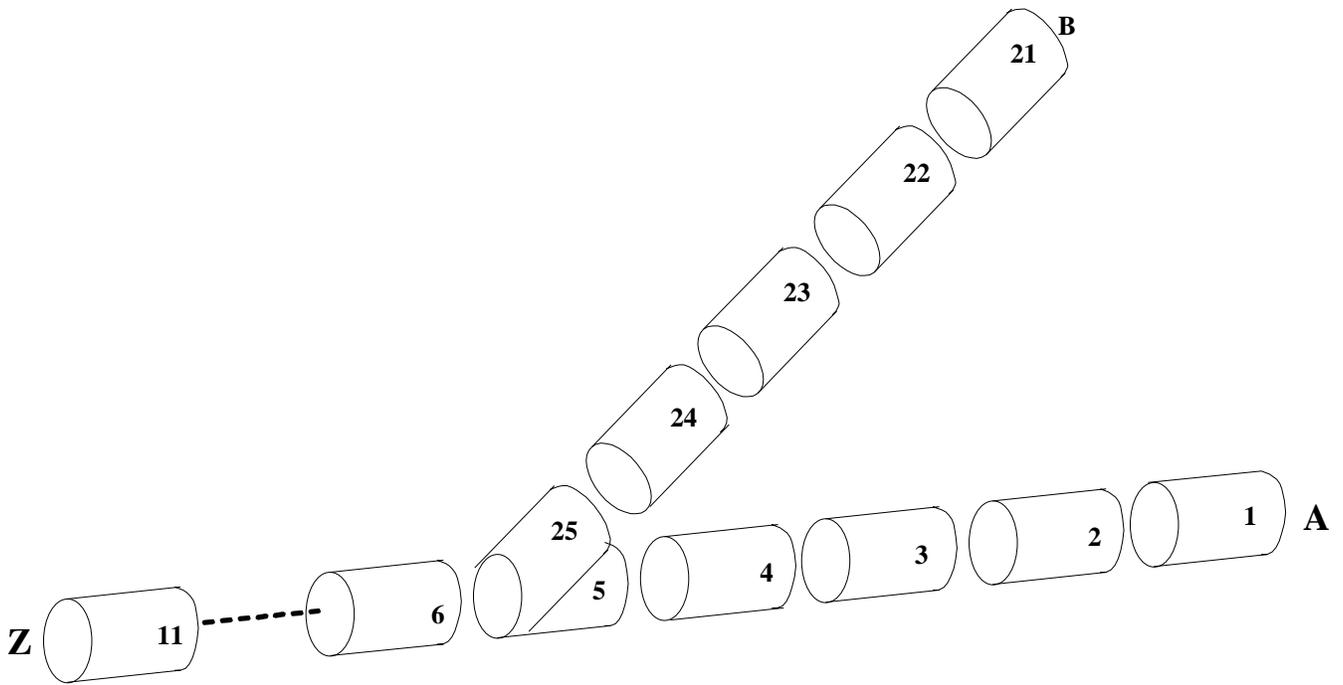

**Figure 10  A junction of paths to generate logic.**

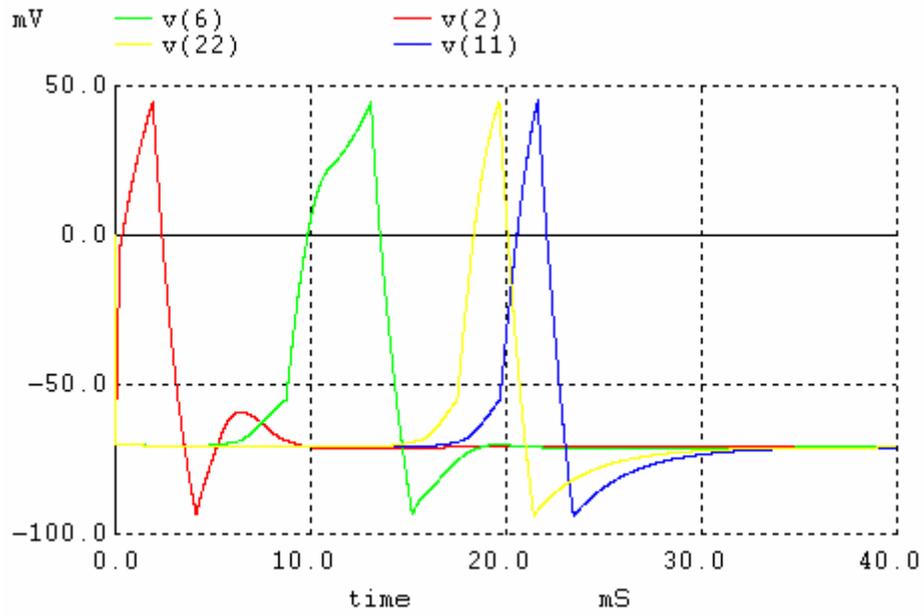

**Figure 11  Only the A segment is triggered, giving output V(11).**

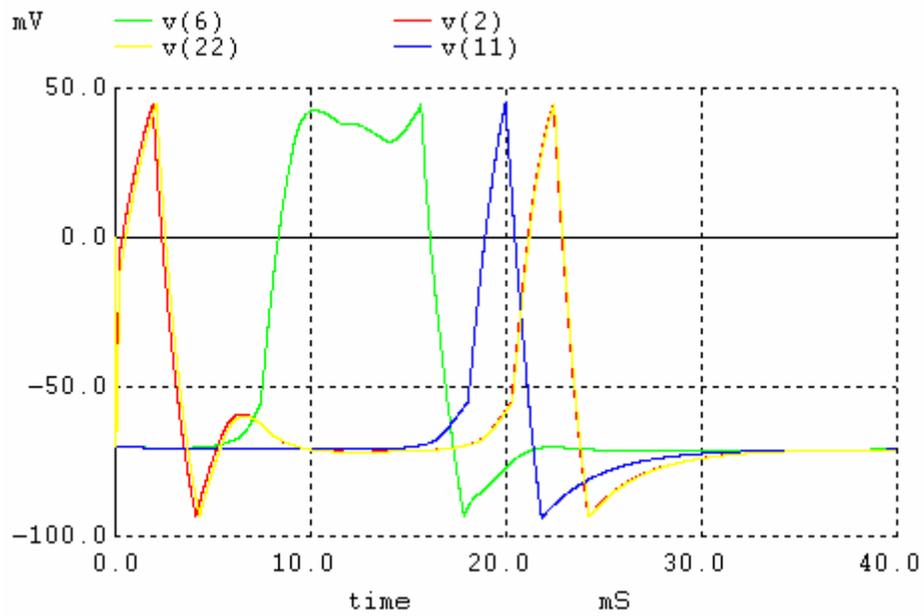

**Figure 12  Both Segments A and C are triggered, giving an output V(11).**

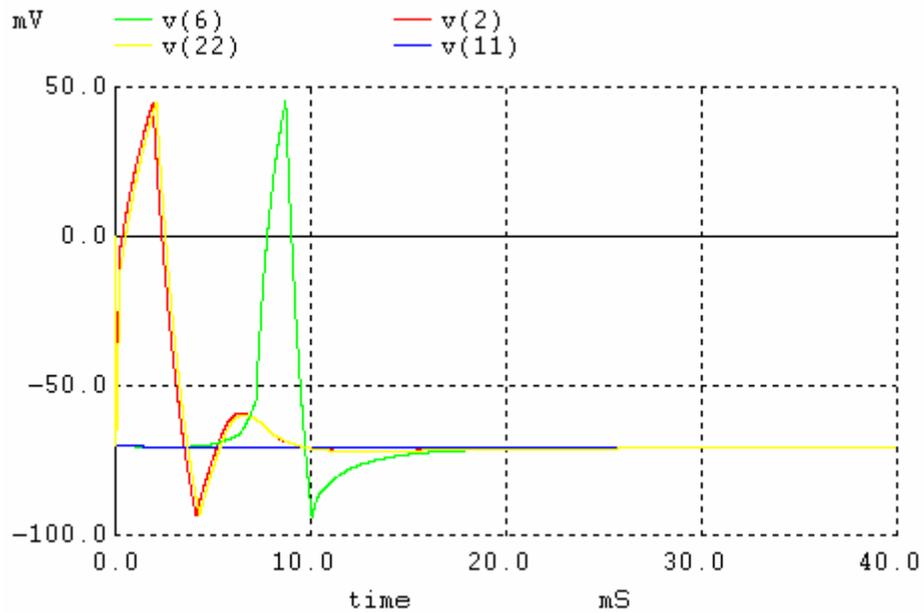

**Figure 13  A and C triggered but no output V(11); XOR because of capacitance reduction in segment 6.**

## AND function

Segment 6 in the above figure may be modified to achieve an AND function: Let L = 0.5e-1 cm. Then $C_1 \approx 15$ pF, $R_1 \approx 100$ M and $R_{LOSS1} \approx 212$ M. The current sources in segment 6 are set to zero to model the effects of inhibitory neurotransmitters. Thus, to propagate a pulse to Z, there must be enough charge to move passively through the shortened segment 6 and to trigger segment 7. The result is that two inputs, one from A and one from B are required to propagate a pulse to Z, as in Figure 14. Triggering only input A, for example, is insufficient to propagate a pulse to Z as shown in Figure 15.

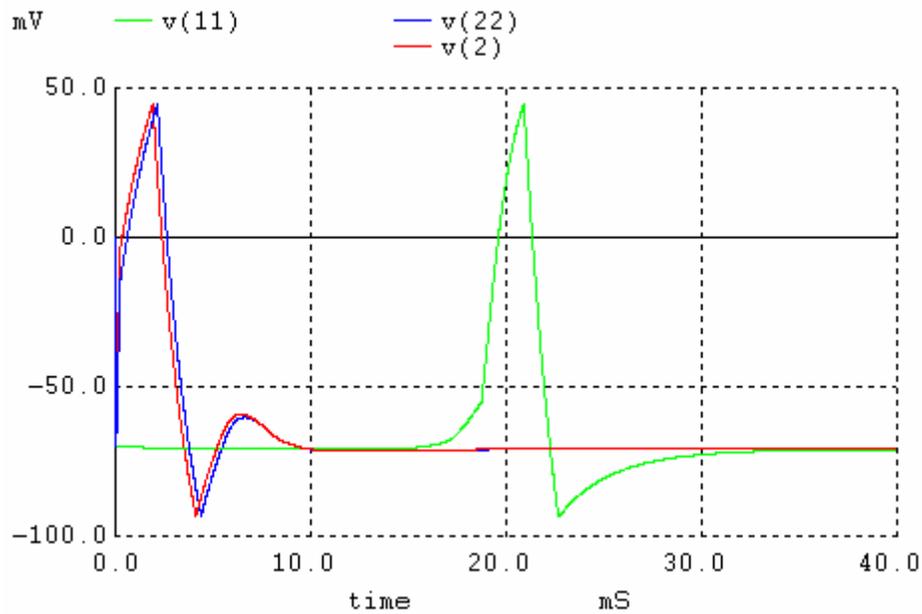

**Figure 14  The AND results by shortening and deactivating segment 6.**

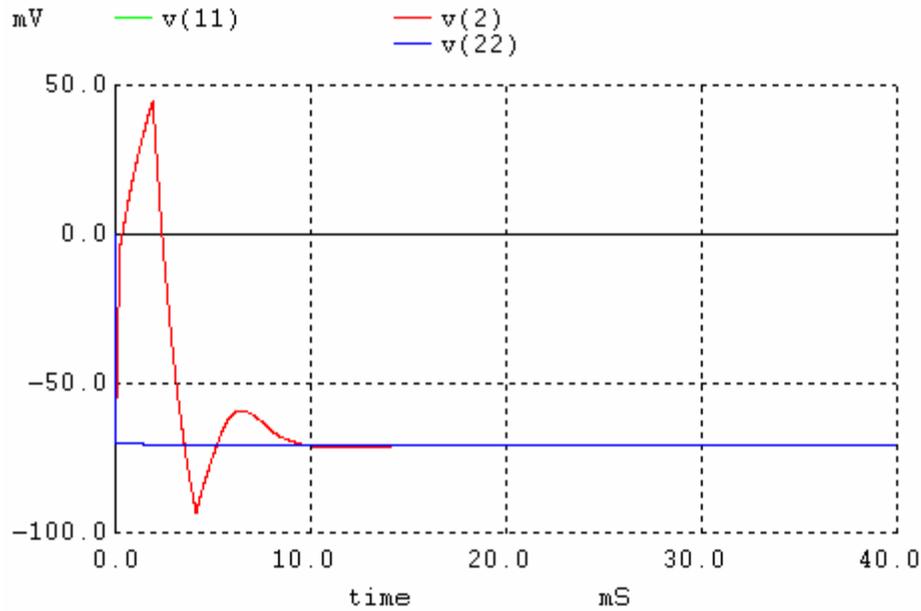

**Figure 15  Only v(2) is present; v(22) is zero; no output at V(11).**

## *Tapered circuits*

A tapered circuit is a tube that begins with a given diameter and ends with a smaller diameter as in Figure 16.  Simulations of tapered circuits suggest that it is slightly easier for a soliton to propagate from the larger end to the smaller end.  This is reasonable because it is easier to charge a slightly smaller capacitor through a slightly smaller resistor, and hence trigger a smaller segment.  Note that if propagation occurs from the smaller end to the larger end, then certainly back propagation is possible from the larger end to the smaller end.

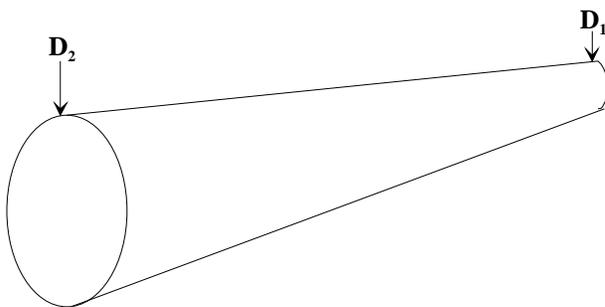

**Figure 16  Tapered circuit.**

Dendritic logic as above produces pulses only when needed.  Otherwise it is in standby mode.

**Modeling Conclusions**

Millions of years of evolution have produced the modern neuron, the basic element of the human brain. Covering the neuron is a ferroelectric membrane, which was modeled with current sources to produce electrical solitons. The propagation properties of electrical solitons were investigated using a standard circuit simulator. In general, solitons in dendrites and axons were found to depend on a judicious balance of resistance and capacitance. An example of a useful balance was simulated above, although other voltage levels and pulse widths are possible.

It was shown that solitons obey arbitrary Boolean logic using fairly obvious assumptions about the roles of excitatory and inhibitory neurotransmitter ions. This style of neural logic points to artificial membranes, and indeed, artificial neural logic is currently being envisioned.

**Historical References**